\documentclass[11pt]{article}
\usepackage{coling2018}
\usepackage{times}
\usepackage{url}
\usepackage{latexsym}
\usepackage{tabularx}

\title{Recognizing Arrow Of Time In The Short Stories}

\author{Fahimeh Hosseini \\
  Shenakht Pajouh / Sharif \\
   University of Technology,\\
    Tehran, Iran \\
  \tt fahim.hosseini.77@ \\
  \tt gmail.com \\\And
  Hosein Fooladi \\
  Shenakht Pajouh / Sharif \\
  University of Technology, \\
  Tehran, Iran \\
  \tt fooladi.hosein@ \\
  \tt gmail.com \\\And
  Mohammad Reza Samsami \\
  Shenakht Pajouh / Sharif \\
  University of Technology, \\
  Tehran, Iran \\
  \tt mohammadrezasamsami76@ \\
  \tt gmail.com \\}

\date{}

\begin{document}
\maketitle
\begin{abstract}
Recognizing arrow of time in short stories is a challenging task. i.e., given only two paragraphs, determining which comes first and which comes next is a difficult task even for humans. In this paper, we have collected and curated a novel dataset for tackling this challenging task. We have shown that a pre-trained BERT architecture achieves reasonable accuracy on the task, and outperforms RNN-based architectures. 
\end{abstract}

\section{Introduction}
\label{intro}

%
%
    %
    %
    %
    %
    %
    %

Recurrent neural networks (RNN) and architectures based on RNNs like LSTM ~\cite{Hochreiter:97} has been used to process sequential data more than a decade. Recently, alternative architectures such as convolutional networks~\cite{Dauphin:17,Gehring:17} and transformer model ~\cite{Vaswani:17} have been used extensively and achieved the state of the art result in diverse natural language processing (NLP) tasks. Specifically, pre-trained models such as the OpenAI transformer~\cite{Radford:18} and BERT~\cite{Devlin:18} which are based on transformer architecture, have significantly improved accuracy on different benchmarks.

In this paper, we are introducing a new dataset which we call \textit{ParagraphOrdering}, and test the ability of the mentioned models on this newly introduced dataset. We have got inspiration from "Learning and Using the Arrow of Time" paper ~\cite{Wei:18} for defining our task. They sought to understand the arrow of time in the videos; Given ordered frames from the video, whether the video is playing backward or forward. They hypothesized that the deep learning algorithm should have the good grasp of the physics principle (e.g. water flows downward) to be able to predict the frame orders in time.

Getting inspiration from this work, we have defined a similar task in the domain of NLP. Given two paragraphs, whether the second paragraph comes really after the first one or the order has been reversed. It is the way of learning the arrow of times in the stories and can be very beneficial in neural story generation tasks. Moreover, this is a self-supervised task, which means the labels come from the text itself. 

\begin{table}[bp]
	\begin{center}
		\begin{tabularx}{\linewidth}{lX}
			\hline Label & 1\\
			\hline First Paragraph & Now they were walking through the trees, one of them carrying him in its huge arms, quite gently. He was scarcely conscious of his surroundings. It was becoming more and more difficult to breathe.\\
			\hline Second Paragraph & Then he felt himself laid down on something soft and dry. The water was not falling on him now. He opened his eyes.\\
			\hline
		\end{tabularx}
	\end{center}
	\caption{\label{Example} A single example of \textit{ParagraphOrdering} dataset.  }
\end{table}

\section{Paragraph Ordering Dataset}

We have prepared a dataset, \textit{ParagraphOrdreing}, which consists of around 300,000 paragraph pairs. We collected our data from Project Gutenberg. We have written an API for gathering and pre-processing in order to have the appropriate format for the defined task.\footnote{API for downloading the dataset:
\url{https://github.com/ShenakhtPajouh/transposition-data}. 
The implementation of different algorithms: \url{https://github.com/ShenakhtPajouh/transposition-simple}}
Each example contains two paragraphs and a label which determines whether the second paragraph comes really after the first paragraph (true order with label 1) or the order has been reversed (Table~\ref{Example}). The detailed statistics of the data can be found in Table~\ref{Statistics}.

\begin{table}[h]
	\begin{center}
		\begin{tabular}{lr}
			\hline \#Train Samples & 294265 \\ 
			\#Test Samples & 32697  \\
			\hline Unique Paragraphs & 239803 \\
			
			\hline Average Number of Tokens & 160.39 \\
			Average Number of Sentences & 9.31\\
			\hline
		\end{tabular}
	\end{center}
	\caption{\label{Statistics} Statistics of \textit{ParagraphOrdering} dataset.  }
\end{table}

\section{Approach}
\label{sec:approach}

Different approaches have been used to solve this task. The best result belongs to classifying order of paragraphs using pre-trained BERT model. It achieves around $84\%$ accuracy on test set which outperforms other models significantly.

\begin{table}[h]
\begin{center}
\begin{tabular}{|l|r|}
\hline \bf Model & \bf Accuracy ($\pm0.01$) \\ \hline
LSTM+Feed-Forward & 0.518  \\
LSTM+Gated CNN+Feed-Forward & 0.524  \\
BERT Features(512 tokens)+Feed-Forward & 0.639  \\
BERT Classifier(30 tokens / 15 tokens from each paragraph) & 0.681 \\
BERT Classifier(128 tokens / 64 tokens from each paragraph) & 0.717 \\
BERT Classifier(256 tokens / 128 tokens from each paragraph) & 0.843 \\
\hline
\end{tabular}
\end{center}
\caption{\label{Accuracy} Accuracy on Test set.  }
\end{table}

\subsection{Encoding with LSTM and Gated CNN}
\label{sect:lstmgated}

In this method, paragraphs are encoded separately, and the concatenation of the resulted encoding is going through the classifier. First, each paragraph is encoded with LSTM. The hidden state at the end of each sentence is extracted, and the resulting matrix is going through gated CNN~\cite{Dauphin:17} for extraction of single encoding for each paragraph. The accuracy is barely above $50\%$, which depicts that this method is not very promising.

\subsection{Fine-tuning BERT}
\label{ssec:BERT}
We have used a pre-trained BERT in two different ways. First, as a feature extractor without fine-tuning, and second, by fine-tuning the weights during training. The classification is completely based on the BERT paper, i.e., we represent the first and second paragraph as a single packed sequence, with the first paragraph using the A embedding and the second paragraph using the B embedding. In the case of feature extraction, the network weights freeze and CLS token are fed to the classifier. In the case of fine-tuning, we have used different numbers for maximum sequence length to test the capability of BERT in this task. First, just the last sentence of the first paragraph and the beginning sentence of the second paragraph has been used for classification. We wanted to know whether two sentences are enough for ordering classification or not. 
After that, we increased the number of tokens and accuracy respectively increases. We found this method very promising and the accuracy significantly increases with respect to previous methods (Table~\ref{Accuracy}). This result reveals fine-tuning pre-trained BERT can approximately learn the order of the paragraphs and arrow of the time in the stories.


\end{document}